\documentclass{ifacconf}
\usepackage{natbib}        
\usepackage{amsfonts,amssymb,amsmath,mathtools,multicol}
\usepackage{physics}
\usepackage{multirow}
\usepackage{float}
\usepackage{graphicx}
\graphicspath{{images/}}

\usepackage{mathrsfs}
\usepackage{subcaption}
\usepackage[dvipsnames]{xcolor}

\usepackage{algorithm}
\usepackage[indLines=true, 
		     noEnd=true, 
		     beginLComment=//~, 
		     endLComment=~, 
		     beginComment=//~]{algpseudocodex}

\usepackage[inline]{enumitem}
\usepackage{cancel}

\usepackage{tabularx}   
\usepackage{booktabs}   
\usepackage{url}

\newcommand{\V}[1]{{\boldsymbol{\mathbf{#1}}}}

\newcommand{\Vdot}[1]{\dot{\V{#1}}}

\begin{document}
\begin{frontmatter}

\title{Whole-body Motion Planning and Safety-Critical Control for Aerial Manipulation} 

\thanks[corrauthor]{Corresponding author$^{\dagger}$}
\thanks[funding]{This work was partly supported by the Institute of Information \& Communications Technology Planning \& Evaluation(IITP)-ITRC(Information Technology Research Center) grant funded by the Korea government(MSIT)(IITP-2025-RS-2024-00437268), and the National Research Foundation, Singapore, under the NRF Medium Sized Centre scheme (CARTIN) and the InnoCORE program of the Ministry of Science and ICT(N10250155).}
\author[First]{Lin Yang} 
\author[Second]{Jinwoo Lee}
\author[First]{Domenico Campolo}
\author[Third]{H. Jin Kim}
\author[Third]{Jeonghyun Byun$^{\dagger}$}
\address[First]{School of Mechanical and Aerospace Engineering, Nanyang Technological University, Singapore. (e-mail: yang0752@e.ntu.edu.sg, d.campolo@ntu.edu.sg)}
\address[Second]{Department of Aerospace Engineering, Seoul National University, South Korea. (e-mail: jinwoolee0728@snu.ac.kr)}
\address[Third]{Automation and Systems Research Institute, Seoul National University, South Korea. (e-mail: \{hjinkim, quswjdgus97\}@snu.ac.kr)}

\begin{abstract}                
Aerial manipulation combines the maneuverability of multirotors with the dexterity of robotic arms to perform complex tasks in cluttered spaces. Yet planning safe, dynamically feasible trajectories remains difficult due to whole-body collision avoidance and the conservativeness of common geometric abstractions such as bounding boxes or ellipsoids. 
We present a whole-body motion planning and safety-critical control framework for aerial manipulators built on superquadrics (SQs). Using an SQ-plus-proxy representation, we model both the vehicle and obstacles with differentiable, geometry-accurate surfaces. Leveraging this representation, we introduce a maximum-clearance planner that fuses Voronoi diagrams with an equilibrium-manifold formulation to generate smooth, collision-aware trajectories. We further design a safety-critical controller that jointly enforces thrust limits and collision avoidance via high-order control barrier functions.
In simulation, our approach outperforms sampling-based planners in cluttered environments, producing faster, safer, and smoother trajectories and exceeding ellipsoid-based baselines in geometric fidelity. Actual experiments on a physical aerial-manipulation platform confirm feasibility and robustness, demonstrating consistent performance across simulation and hardware settings. The video can be found at \url{https://youtu.be/hQYKwrWf1Ak}.
\vspace{-1mm}
\end{abstract}


\end{frontmatter}

\section{Introduction}
\vspace{-1mm}

\begin{figure}[t]
\centering
\includegraphics[width = 0.36\textwidth]{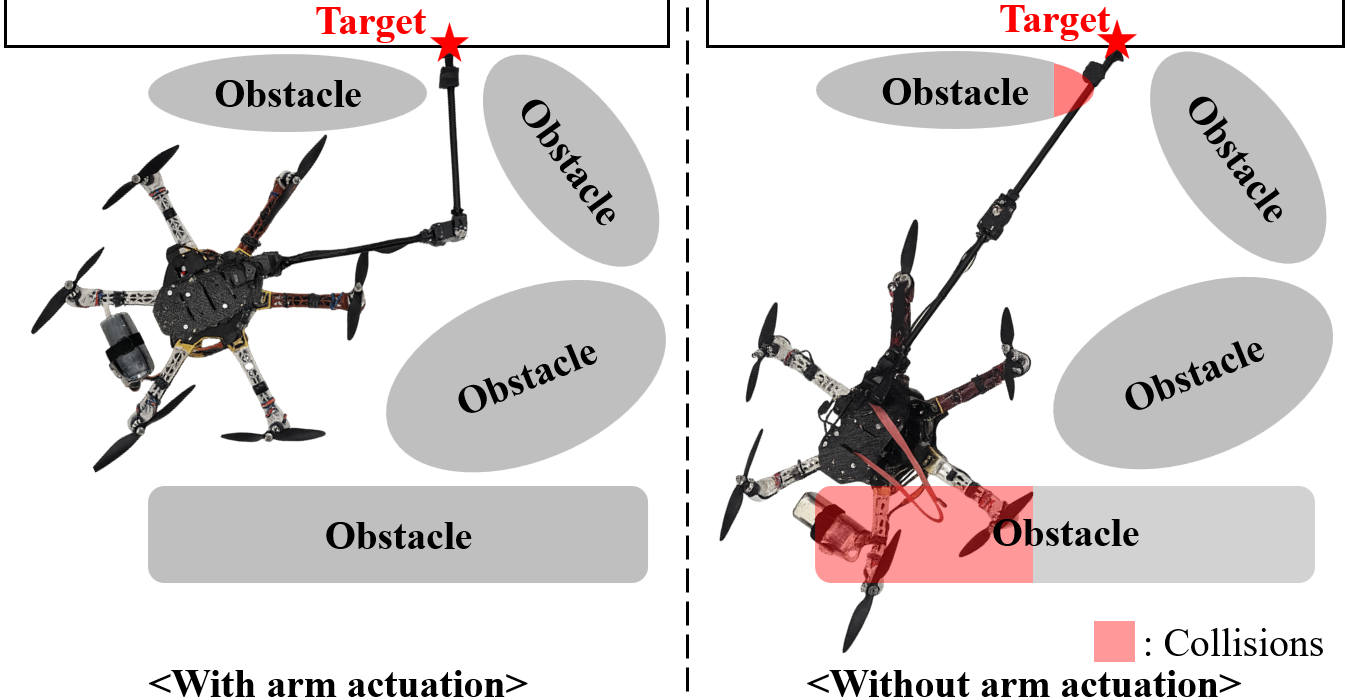}
\vspace{-0.36cm}
\caption{Collision avoidance in a cluttered environment.} \label{fig: motivation}
\end{figure}

Aerial manipulation combines the maneuverability of aerial vehicles with the dexterity of robot arms. Leveraging this synergy, numerous studies have been conducted on diverse applications, including door opening, plug-pulling and non-destructive testing (NDT). Bringing such applications into real-world operation requires prioritizing safety --most notably, preventing unintended collisions. This is especially important for aerial manipulators, whose end effectors can reach targets in tight or occluded spaces that would be inaccessible without arm actuation, thereby increasing both opportunity and collision risk as shown in Fig. \ref{fig: motivation}. Furthermore, in the control level, both dynamic feasibility and instantaneous collision avoidance must be satisfied, as the vehicle may be unable to follow the desired trajectory generated from planners due to the limitations in control performances. However, very few studies have simultaneously addressed collision avoidance between the arm’s linkages and surrounding obstacles and the system's dynamic feasibility. 
\vspace{-1mm}

\vspace{-1mm}

\subsection{Related Works}
\vspace{-1mm}

There have been several studies on collision avoidance of aerial manipulators (AMs). \cite{seo2017locally} presents a locally optimal trajectory planning method for aerial manipulation in constrained environments, and \cite{cao2024motion} proposes a motion planner for aerial pick-and-place tasks that incorporates collision avoidance constraints between the movable range of the robot arm and surrounding obstacles. Additionally, \cite{zhang2025end} introduces an end-effector-oriented collision avoiding motion planner for AMs accounting for system dynamics. However, these methods only focus on the poses of the multirotor and the end-effector, not the collisions between the robot arm’s linkages and obstacles.
\vspace{-1mm}

To prevent possible collisions between the robot arm's linkages and the surrounding environment, there exist other studies that consider such collisions. \cite{liu2024coordinated} presents a model predictive control framework employing simple box constraints on the states of both the multirotor and the robot arm, and \cite{lee2025autonomous} proposes a whole-body planner for omnidirectional AMs, where both the multirotor and the robot arm’s linkages are represented as ellipsoids. Despite these advances, such approaches rely on bounding box or ellipsoid-based geometric representations for AMs and obstacles, which increase conservativeness when navigating through narrow passages.
\vspace{-1mm}

To overcome the conservative issue, several studies have developed sampling-based methods for collision avoidance. In \cite{yavari2022interleaved}, the authors present a rapidly-exploring random trees star (RRT*)-based planner for collision avoidance. However, sampling-based methods are generally more time-consuming than approaches based on geometric representations.
\vspace{-1mm}

\subsection{Contributions}
\vspace{-1mm}

In this paper, we propose a whole-body motion planning and a safety-critical control of an AM using superquadrics and proxies, wherein both the multirotor platform and the robot arm's linkages independently achieve collision avoidance with surrounding obstacles. To the best of the authors’ knowledge, this work is the first to employ a superquadric–proxy representation for AMs, enabling accurate yet efficient whole-body modeling and differentiable distance evaluation for collision avoidance. Also, we design a safety-critical framework that integrates the SQ–proxy representation with Voronoi-based maximum-clearance planning and control barrier functions (CBFs), ensuring dynamic feasibility and instantaneous collision avoidance under thrust limits. Furthermore, comparative simulations and real-world experiments demonstrate that our approach achieves faster, safer, and smoother trajectories than sampling-based planners, while surpassing ellipsoid-based baselines in geometry accuracy.
\vspace{-1mm}

\subsection{Notations}
\vspace{-1mm}

$\boldsymbol{0_{i \times j}}$, $\boldsymbol{I_{i}}$, and $\boldsymbol{e_3}$ denote the $i\times j$ zero matrix, $i \times i$ identity matrix, and unit vector along the $z$-axis, respectively; for scalars $a_1,\cdots,a_N$, we use $c a_1$ and $s a_1$ for $\cos a_1$ and $\sin a_1$, and diag$\{a_1,\cdots,a_N\}$ for the $N \times N$ diagonal matrix with $a_i$ on the $(i,i)$-th entry; for vectors $\boldsymbol{\alpha}, \boldsymbol{\beta}$, $\alpha_i$ denotes the $i$-th element of $\boldsymbol{\alpha}$, and if $\boldsymbol{\alpha}, \boldsymbol{\beta} \in \mathbb{R}^3$, then $[\boldsymbol{\alpha}]_{\times} \in \mathbb{R}^{3\times3}$ is the skew-symmetric matrix satisfying $[\boldsymbol{\alpha}]_{\times}\boldsymbol{\beta} = \boldsymbol{\alpha} \times \boldsymbol{\beta}$; for matrices $\boldsymbol{A_1},\cdots,\boldsymbol{A_N}$, $\textrm{blkdiag}\{\boldsymbol{A_1},\cdots,\boldsymbol{A_N}\}$ denotes the block diagonal matrix formed by aligning them; and "with respect to" is abbreviated as w.r.t.
\vspace{-1mm}

\section{Aerial Manipulator System}
\vspace{-1mm}

\begin{figure}[t]
\centering
\vspace{0.00cm}
\includegraphics[width = 0.32\textwidth]{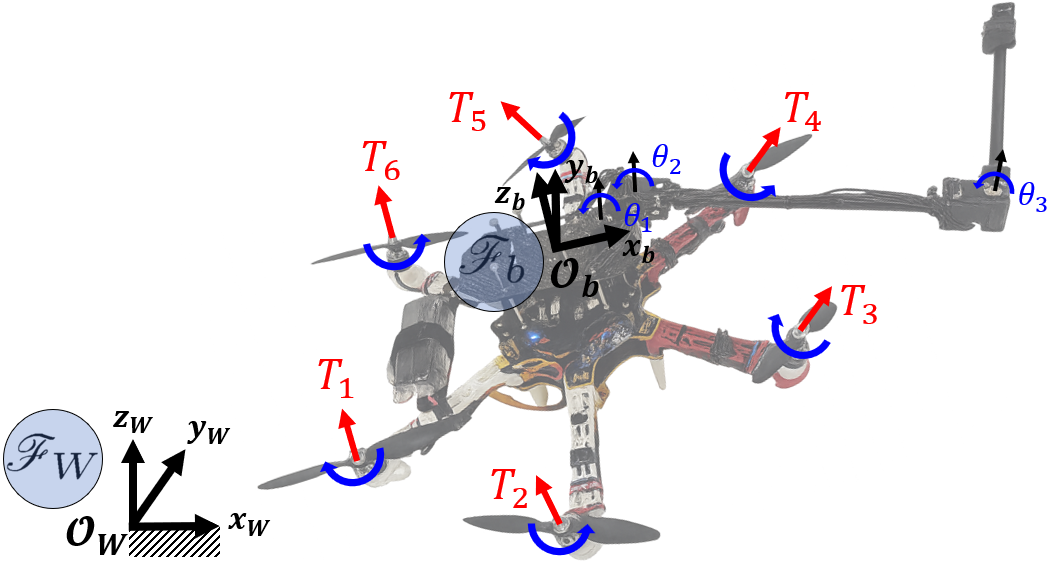}
\vspace{-0.40cm}
\caption{An AM configured with a fully actuated multirotor and a 3-DOF robot arm with the thrust vector, $\boldsymbol{T} \in {\mathbb{R}}^{6}$, the joint angles of the robot arm, $[\theta_1;\theta_2;\theta_3]$, and the Earth-fixed and multirotor frames, $\mathscr{F}_{W}$ and $\mathscr{F}_b$.} \label{fig: aerial manipulator with fully actuated multirotor}
\end{figure}

We utilize an AM configured with a fully actuated multirotor with six tilted motors and a 3-DOF robot arm shown in Fig. \ref{fig: aerial manipulator with fully actuated multirotor}. We let $\mathscr{F}_W$ and $\mathscr{F}_b$ denote two key coordinate frames, the Earth-fixed and multirotor body frames, respectively, then the position and ZYX Euler angles of $\mathscr{F}_b$ w.r.t. $\mathscr{F}_W$ are defined as $\boldsymbol{p}$ and $\boldsymbol{\phi} \in {\mathbb{R}}^{3}$, respectively. Since we consider the movement of the robot arm as a part of external disturbances, the generalized coordinate of the AM, $\boldsymbol{q}$ $\triangleq [\boldsymbol{p};\boldsymbol{\phi}]$.
\vspace{-1mm}

\vspace{-1mm}

According to \cite{byun2025safety}, the Euler-Lagrange model of the AM is formulated as follows: 
\begin{equation} \label{eqn: dynamic model of the aerial manipulator}
    \begin{split}
       \boldsymbol{M}(\boldsymbol{\phi})\boldsymbol{\ddot{q}} + \boldsymbol{C}(\boldsymbol{\phi},\boldsymbol{\dot{\phi}}) + \boldsymbol{G}  =  \boldsymbol{\tau} + \boldsymbol{\tau_{ext}}
    \end{split}
\end{equation}
where $\boldsymbol{\tau}$ and $\boldsymbol{\tau_{ext}} \in {\mathbb{R}}^6$ represent the generalized control wrench and external disturbances, respectively, and 
\begin{equation*}
    \begin{split}
        \boldsymbol{M}(\boldsymbol{\phi}) \triangleq& \begin{bmatrix}
            m\boldsymbol{I_3} & \boldsymbol{0_{3\times3}} \\
            \boldsymbol{0_{3\times3}} & \boldsymbol{Q^{\top}}\boldsymbol{J}\boldsymbol{Q}
        \end{bmatrix}, \\
        \boldsymbol{C} (\boldsymbol{\phi},\boldsymbol{\dot{\phi}}) \triangleq& \begin{bmatrix}
            \boldsymbol{0_{3\times1}} \\
            \boldsymbol{Q^{\top}}(\boldsymbol{J}\boldsymbol{\dot{Q}}\boldsymbol{\dot{\phi}} + [\boldsymbol{Q}\boldsymbol{\dot{\phi}}]_{\times}\boldsymbol{J}\boldsymbol{Q}\boldsymbol{\dot{\phi}})
        \end{bmatrix}, \ \boldsymbol{G} \triangleq \begin{bmatrix}
            mg\boldsymbol{e_3} \\ 
            \boldsymbol{0_{3\times1}}
        \end{bmatrix}
    \end{split}
\end{equation*}
with the mass and moment of inertia of the AM, $m$ and $\boldsymbol{J} \in {\mathbb{R}}^{3\times3}$, and the gravitational acceleration, $g$. Also, we let $\boldsymbol{Q} \in {\mathbb{R}}^{3\times3}$ denote the mapping matrix satisfying $\boldsymbol{\omega} = \boldsymbol{Q}\boldsymbol{\dot{\phi}}$ where $\boldsymbol{\omega} \in {\mathbb{R}}^{3}$ represents the angular velocity of the vehicle w.r.t. $\mathscr{F}_{W}$ expressed in $\mathscr{F}_{b}$. Meanwhile, $\boldsymbol{\tau}$ and $\boldsymbol{T} \triangleq [T_1;\cdots;T_6]$ satisfy $\boldsymbol{\tau} = \boldsymbol{B}(\boldsymbol{\phi})\boldsymbol{T}$ where $\boldsymbol{B}(\boldsymbol{\phi}) \triangleq \textrm{blkdiag}\{\boldsymbol{R}, \boldsymbol{Q^{\top}}\}\boldsymbol{\Xi}(L,\alpha_p,k_f)$ (defined in \cite{byun2025safety}) with the length from the vehicle's origin to each propeller, $L = 0.278$ [m], fixed tilt angle of each motor $\alpha_p = 15 ^{\circ}$ and thrust-to-torque coefficient, $k_f = 0.016$ [m].  
\vspace{-1mm}

To address model uncertainties, (\ref{eqn: dynamic model of the aerial manipulator}) is rearranged as:
\begin{equation} \label{eqn: rearranged dynamic model of the aerial manipulator}
    \begin{split}
       \boldsymbol{\hat{M}}(\boldsymbol{\phi})\boldsymbol{\ddot{q}} + \boldsymbol{\hat{C}}(\boldsymbol{\phi},\boldsymbol{\dot{\phi}}) + \boldsymbol{\hat{G}} = \boldsymbol{\tau} + \boldsymbol{d}
    \end{split}
\end{equation}
where $\boldsymbol{\hat{M}}(\boldsymbol{\phi})$, $\boldsymbol{\hat{C}}(\boldsymbol{\phi},\boldsymbol{\dot{\phi}})$ and $\boldsymbol{\hat{G}}$ are the nominal values of $\boldsymbol{M}(\boldsymbol{\phi})$, $\boldsymbol{C}(\boldsymbol{\phi},\boldsymbol{\dot{\phi}})$ and $\boldsymbol{G}$, respectively, and the lumped disturbance $\boldsymbol{d} \in {\mathbb{R}}^6$ is derived as follows: 
\begin{equation*} 
    \boldsymbol{d} \triangleq (\boldsymbol{\hat{M}}(\boldsymbol{\phi})-\boldsymbol{M}(\boldsymbol{\phi}))\boldsymbol{\ddot{q}} + \boldsymbol{\hat{C}}(\boldsymbol{\phi},\boldsymbol{\dot{\phi}}) - \boldsymbol{C}(\boldsymbol{\phi},\boldsymbol{\dot{\phi}}) + \boldsymbol{\hat{G}} - \boldsymbol{G} + \boldsymbol{\tau_{ext}}.
\end{equation*}
\vspace{-1mm}

\section{Whole-body motion planning}
\vspace{-1mm}

\begin{figure}[t!]
	\centering
	\includegraphics[width=0.70\columnwidth]{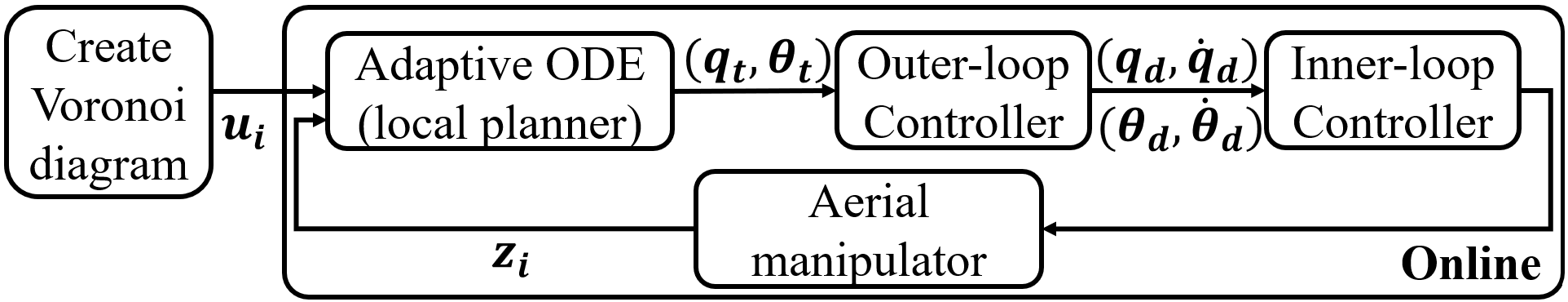}
	\vspace{-0.3cm}
    \caption{Planning and control of AM.}
	\label{FIG: Diagram}
\end{figure}

Building upon \cite{yang2024path,yang2025planning} and \cite{kana2021human}, we model the entire aerial manipulation system as well as environmental obstacles using SQs. The interactions between any two SQs are represented by proxies. Meanwhile, the end-effector of the robot arm follows Voronoi edges, while collision avoidance is also handled via proxies. The entire pipeline is illustrated in Fig. \ref{FIG: Diagram}.
\vspace{-1mm}

\vspace{-1mm}

\subsection{Superquadrics and proxies}
\vspace{-1mm}

Following \cite{jaklic2000segmentation,yang2025planning}, we represent robot bodies and obstacles by SQs:
\begin{equation}\label{eq:SQ_3D}
    \begin{split}
        F_{3d}(x,y,z) = \Big((\tfrac{x}{a_{1}})^{\tfrac{2}{\epsilon_2}} + (\tfrac{y}{a_{2}})^{\tfrac{2}{\epsilon_2}}\Big)^{\tfrac{\epsilon_2}{\epsilon_1}} + (\tfrac{z}{a_{3}})^{\tfrac{2}{\epsilon_1}} - 1
    \end{split}
\end{equation}
where $a_i$ and $\epsilon_i$ determine the size and shape, respectively. A proxy point on the SQ boundary is parameterized by
\begin{align}\label{eqn: 3D proxy} 
        \V p (\boldsymbol{\gamma_{3D}}) =  \begin{bmatrix}
    a_1 \cos^{\epsilon_1} \gamma_{3D,1} \cos^{\epsilon_2} \gamma_{3D,2} \\
    a_1 \cos^{\epsilon_1} \gamma_{3D,1} \sin^{\epsilon_2} \gamma_{3D,2}  \\
    a_3 \sin^{\epsilon_1} \gamma_{3D,1} 
    \end{bmatrix}, \nonumber\\
    \boldsymbol{\gamma_{3D}} \in [-\tfrac{\pi}{2}, \ \tfrac{\pi}{2}] \times [-\pi, \pi] 
\end{align} 
where trigonometric powers are signed powers. As excessive tilting is restricted, the planning problem is reduced to its 2D SQ with parameters $a_1,a_2,\epsilon$ and $\gamma\in[-\pi,\pi]$.
\vspace{-1mm}

\subsection{Maximum-clearance path planning}
\vspace{-1mm}

We construct a Voronoi-based planner to maximize obstacle clearance \cite{arslan2019sensor}, following the SQ-proxy formulation in \cite{yang2024path}. Given disjoint obstacle SQs, the boundary between two Voronoi cells is defined by the maximum-margin separating hyperplane:
\begin{equation*}
\begin{split}
    HP_{i,j} &:= \left\{ \V p \in E \ \middle|\ 
    \|\V p-\V p(\gamma_i)\|=\|\V p-\V p(\gamma_j)\| \right\},\\
    V_i &:= \left\{ \V p \in W \ \middle|\ 
    \|\V p-\V p(\gamma_i)\| \leq \|\V p-\V p(\gamma_j)\|,\ \forall j\neq i \right\},
\end{split}
\end{equation*}
where the proxy points $\V p(\gamma_i)$ and $\V p(\gamma_j)$ are computed from Eq. \ref{eqn: 3D proxy}. Moving along $HP_{i,j}$ keeps the path equidistant from the corresponding obstacles, yielding a maximum-clearance path. Within the bounded workspace $W$, since each Voronoi cell is obtained from the intersection of its half-space constraints, $V_i = \bigcap_{j=0}^{m} W \cap HP_{i,j}$, the Voronoi diagram is converted into a weighted graph $\mathcal{G}=(\mathcal{V},\mathcal{E})$ where vertices are graph nodes and edge weights are Euclidean distances. A graph search then returns the global solution path $(\mathcal{V}_{\mathrm{sol}}, \mathcal{E}_{\mathrm{sol}})$.
\vspace{-1mm}

\subsection{Planning on equilibrium manifold}
\vspace{-1mm}

Following \cite{yang2025planning,campolo2025geometric}, we formulate planning through a manipulation potential $W:\mathcal{Z}\times\mathcal{U}\rightarrow\mathbb{R}$, where $\V z$ is the aerial-manipulator state and $\V u$ is the end-effector attracting pose. A smooth collision-avoiding trajectory is obtained by tracking the equilibrium state $\V z^*$ on the equilibrium manifold (EM):
\begin{equation}
\mathcal{M}^{eq}:=\{(\V z,\V u)\in\mathcal{Z}\times\mathcal{U}\mid
\partial_{\V z}W(\V z,\V u)=\V 0\}.
\end{equation}
Since $\V z^*$ is implicitly defined, we update it using the adaptive ODE in \cite{yang2025planning}:
\begin{equation}
\Vdot z =
-(\partial^2_{\V{zz}}W)^{-1}\partial^2_{\V{uz}}W\Vdot u
-\eta(\partial^2_{\V{zz}}W)^{-1}\partial_{\V z}W ,
\label{eq:adaptive ODE}
\end{equation}
where $\eta>0$ is the integration gain.
\vspace{-1mm}

\subsection{System modeling}
\vspace{-1mm}


\begin{figure}[t!]
	\centering
	\includegraphics[width=0.45\columnwidth]{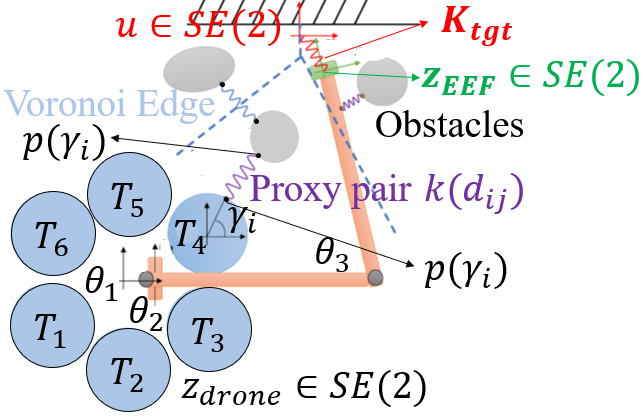}
    \vspace{-0.35cm}
	\caption{System representation using SQs and proxies.}
	\label{FIG:System_Model}
\end{figure}
\vspace{-1mm}

As shown in Fig. \ref{FIG:System_Model}, the AM is modeled in 2D with $\V z_{mr}\in SE(2)$ and two active arm joints $\theta_1,\theta_3$. The six propellers and two arm links are represented by eight SQs, $SQ_i,\ i=1,\ldots,8$. For each robot SQ and obstacle SQ, a proxy pair is assigned, yielding the system state
$\V z=[\V z_{sys},\Gamma]$, where $\V z_{sys}=[\V z_{mr},\theta_1,\theta_3]$ and $\Gamma$ collects all proxy variables. The end-effector pose $\V z_{EEF}\in SE(2)$ is obtained by forward kinematics.
\vspace{-1mm}

\subsubsection{(1)}
Collision avoidance is encoded by the proxy potential $W_{px}=\sum_i \sum_j \frac{1}{2} k(d_{ij}-d') \|\V p(\gamma_i) - \V p(\gamma_j) \|^2$ where $d_{ij}=F({}^j\V p(\gamma_i))$ evaluates the proxy of robot $SQ_i$ in the frame of obstacle $SQ_j$, and $d'$ is the safety margin. The nonlinear stiffness $k(d) = k_\text{min} + \frac{1-\tanh(d/d_0)}{2} k_\text{max}$ increases near obstacles and decreases away from them.

\subsubsection{(2)}
Target following: The attracting pose $\V u=[u_x,u_y,u_\theta]$ $\in SE(2)$ follows $\mathcal{E}_{sol}$ in position and aligns its orientation with the Voronoi-edge normal. The target potential is
\begin{equation}
W_{tgt}= \frac{1}{2} (\V u-\V z_{EEF})^T \V K_{tgt} (\V u-\V z_{EEF}).
\end{equation}
The total manipulation potential is then $W=W_{px}+W_{tgt}$.
\vspace{-1mm}

\subsection{Local planner and Target Pose Calculation}
\vspace{-1mm}

The global planner provides discrete nodes and edges $(\mathcal{V}_{sol},\mathcal{E}_{sol})$, from which each via point is converted into an attraction pose $\V u_i$. To obtain a smooth reference trajectory $\V z^*(s)$, $s\in[0,1]$, we integrate the augmented ODE:
\begin{subequations}
\begin{align}
\tfrac{\partial \V u}{\partial s} &= \V u_{i+1}-\V u_i, \label{eq:ode1}\\
\tfrac{\partial \V z_{sys}}{\partial s} &=
-(\partial^2_{\V{zz}}W)^{-1}\partial^2_{\V{uz}}W
\tfrac{\partial \V u}{\partial s}
-\eta(\partial^2_{\V{zz}}W)^{-1}\partial_{\V z}W, \label{eq:ode2}\\
\tfrac{\partial \V\Gamma}{\partial s} &= -\alpha\partial_{\V\gamma}W, \label{eq:ode3}
\end{align}
\label{eq:allODE}
\end{subequations}
initialized from the current aerial-manipulator state.
\vspace{-1mm}

From the results of the local planner, the target pose of the multirotor, $\boldsymbol{q_t}$, and the target angles of the robot arm's servo motors, $\boldsymbol{\theta_t}$, are calculated as follows:
\begin{equation} \label{target pose calculation}
    \begin{split}
        \boldsymbol{q_{t}} =& [z_{sys,1}(s(t));z_{sys,2}(s(t));h_{t};0;0;z_{sys,3}(s(t))] \\
        \boldsymbol{\theta_t} =& [z_{sys,4}(s(t));0;z_{sys,5}(s(t))]
    \end{split}
\end{equation}
where the time allocated path parameter $s(t)$ is calculated as $\tfrac{1}{T_d}t$ with the desired task execution time $T_d > 0$, and $h_t$ represents the target altitude of the multirotor.
\vspace{-1mm}

\section{Safety-Critical Controller}
\vspace{-1mm}

This section presents a safety-critical controller subject to two safety constraints. The first constraint enforces propeller-thrust limits: when $\boldsymbol{q_t}$ is substantially far from the current pose $\boldsymbol{q}$, the required control effort can drive one or more thrust beyond their physical bounds. The second constraint enforces instantaneous collision avoidance. This addresses cases where tracking errors between $(\boldsymbol{q}, \boldsymbol{\theta})$ and $(\boldsymbol{q_t}, \boldsymbol{\theta_t})$ arises due to limited control authority, even when the planned global path is collision-free. Our controller is divided into outer- and inner-loop control laws.
\vspace{-1mm}

\subsection{Inner-Loop Control Law}
\vspace{-1mm}

To cope with three performance deteriorating factors, i.e., movement of the attached robot arm, model uncertainties, and time-varying external disturbance, we adopt a DOB-based controller extending the control law introduced in \cite{kim2017robust} to incorporate 6 DOF pose.
\vspace{-1mm}

Let $\boldsymbol{q_d}$ and $\boldsymbol{\theta_d}$ denote the desired pose of the multirotor and the desired joint angles of the robot arm, respectively, which the multirotor and robot arm attempt to track. However, these may differ from the target values, $\boldsymbol{q_t}$ and $\boldsymbol{\theta_t}$, since an \textit{outer-loop controller} is also designed to adjust the target values in accordance with the safety constraints. This controller will be introduced in a later subsection.
\vspace{-1mm}

With $\boldsymbol{q_d} \ \& \ \boldsymbol{\theta_d}$, our inner-loop controller is formulated as:
\begin{equation} \label{eqn: DOB-based controller}
    \begin{split}
        \boldsymbol{\tau} = \boldsymbol{\hat{M}}(\boldsymbol{\phi})(\boldsymbol{K_d}\boldsymbol{\dot{e}} + \boldsymbol{K_p}\boldsymbol{e}) + \boldsymbol{\hat{C}}(\boldsymbol{\phi},\boldsymbol{\dot{\phi}}) + \boldsymbol{\hat{G}} - \boldsymbol{\hat{d}} \\
        \boldsymbol{\hat{d}} = -\boldsymbol{\hat{M}}(\boldsymbol{\phi})(\boldsymbol{E_1}\boldsymbol{p_{dob}}-\boldsymbol{E_2}\boldsymbol{\dot{q}_{dob}}) + \boldsymbol{\hat{C}}(\boldsymbol{\phi},\boldsymbol{\dot{\phi}}) + \boldsymbol{\hat{G}} \\ 
        \boldsymbol{\dot{q}_{dob}} = \boldsymbol{A_{dob}}\boldsymbol{q_{dob}} + \boldsymbol{B_{dob}}\boldsymbol{q}, \ \boldsymbol{T} = \boldsymbol{B^{-1}}(\boldsymbol{\phi})\boldsymbol{\tau} \\ 
        \boldsymbol{\dot{p}_{dob}} = \boldsymbol{A_{dob}}\boldsymbol{p_{dob}} + \boldsymbol{B_{dob}}\boldsymbol{M^{-1}(\boldsymbol{\phi})}\boldsymbol{\tau}
    \end{split}
\end{equation}
where $\boldsymbol{e} \triangleq \boldsymbol{q_d} - \boldsymbol{q}$, $\boldsymbol{E_1} \triangleq \begin{bmatrix}
    \boldsymbol{I_6} & \boldsymbol{0_{6\times6}}
\end{bmatrix}$ and $\boldsymbol{E_2} \triangleq \begin{bmatrix}
    \boldsymbol{0_{6\times6}} & \boldsymbol{I_6}
\end{bmatrix}$ with user-defined matrices $\boldsymbol{K_p}, \boldsymbol{K_d} \in \mathbb{R}^{6 \times 6}_{>0}$. Also, $\boldsymbol{\hat{d}}$ represents the estimated lumped disturbance, and
\begin{equation*}
    \begin{split}
        \boldsymbol{A_{dob}} \triangleq& \begin{bmatrix}
            \boldsymbol{0_{6\times6}} & \boldsymbol{I_6} \\
            -\boldsymbol{\varepsilon_{dob}^{-2}}\boldsymbol{a_{0}} & -\boldsymbol{\varepsilon_{dob}^{-1}}\boldsymbol{a_{1}} 
        \end{bmatrix}, \quad \boldsymbol{B_{dob}} \triangleq \begin{bmatrix}
            \boldsymbol{0_{6\times6}} \\ 
            \boldsymbol{\varepsilon_{dob}^{-2}}\boldsymbol{a_{0}}
        \end{bmatrix}
    \end{split}
\end{equation*}
with user-defined matrices $\boldsymbol{a_j} \triangleq \textrm{diag}\{a_{j,1},\cdots,a_{j,6}\}$ ($j = 0, 1$) and $\boldsymbol{\varepsilon_{dob}} \triangleq \textrm{diag}\{\varepsilon_{dob,1},\cdots, \varepsilon_{dob,6}\}$ under $0 < a_{0,i}, a_{1,i}$, $\tfrac{1}{2} < \tfrac{a_{0,i}}{a^2_{1,i}}$ and $0 < \varepsilon_{dob,i} < 1$ $\forall i = 1,\cdots,6$.
\vspace{-1mm}

\subsection{Safety Constraints}
\vspace{-1mm}

The two safety conditions are formulated as follows:
\vspace{-1mm}

\subsubsection{(1) Thrust Limit}
\vspace{-1mm}

The thrust values have their lower and upper limits, $\underline{T}$ and $\bar{T}$, as follows:
\begin{equation} \label{eqn: lower and upper limits of propeller thrusts}
    \begin{split}
        \underline{T}\boldsymbol{1_{6}} \leq \boldsymbol{T} \leq \bar{T}\boldsymbol{1_{6}}
    \end{split}
\end{equation}
where $\boldsymbol{1_6} \triangleq [1 \ 1 \ 1 \ 1 \ 1 \ 1]^{\top}$. If one of the propellers reaches its limit, the multirotor may lose controllability.
\vspace{-1mm}

\subsubsection{(2) Collision Avoidance}
\vspace{-1mm}

To compensate for possible target-tracking errors in the proposed inner-loop controller, an additional strategy is necessary for collision avoidance, depending on the AM’s pose and the robot arm’s joint angles in real time.
\vspace{-1mm}

To derive a mathematical expression for this constraint, we utilize superquadrics as defined in Eq.\ref{eq:SQ_3D}.
For $SQ_{i}$ (one of the \textit{vehicle} SQs) and $SQ_{j}$ (one of the \textit{obstacle} SQs) ($i = 1. \cdots, 8$, $j = 1. \cdots, m$), the closest points follows Eq. \ref{eqn: 3D proxy}. 
We let $\boldsymbol{\Delta X} \in \mathbb{R}^{3}$ denote the displacement from the center of $SQ_{j}$ to the proxy of $SQ_{i}$ as follows:
\begin{equation} \label{eqn: Delta X}
    \begin{split}
         \underset{\triangleq [\Delta x; \Delta y; \Delta z]}{\underline{\boldsymbol{\Delta X}}}  \triangleq \boldsymbol{R^{\top}_{j}}(\boldsymbol{p_{i}} - \boldsymbol{p_{j}}) + \boldsymbol{R^{\top}_{j}} \boldsymbol{R_{i}} \boldsymbol{p}(\boldsymbol{\gamma_{3D,i}})
    \end{split}
\end{equation}
where $\boldsymbol{R_i}$ and $\boldsymbol{R_j} \in SO(3)$ represent the rotation matrices from $SQ_i$ to $\mathcal{F}_{W}$ and from $SQ_j$ to $\mathcal{F}_{W}$, respectively,  $\boldsymbol{p_i}$ and $\boldsymbol{p_j} \in \mathbb{R}^3$ mean the displacement from $\mathcal{F}_{W}$ to the centers of $SQ_i$ and $SQ_j$, respectively, and $\boldsymbol{\gamma_{3D,i}}$ is the angular variable of $SQ_i$.
Then, a control barrier function (CBF) for collision avoidance is defined as follows:
\begin{equation} \label{eqn: HOCBF}
    \begin{split}
        h_{co} \triangleq \ln(\Big((\tfrac{\Delta x}{^{j}a_{1}})^{\tfrac{2}{^{j}\epsilon_{2}}} + (\tfrac{\Delta y}{^{j}a_{2}})^{\tfrac{2}{^{j}\epsilon_{2}}}\Big)^{\tfrac{^{j}\epsilon_{2}}{^{j}\epsilon_{1}}} + (\tfrac{\Delta z}{^{j}a_{3}})^{\tfrac{2}{^{j}\epsilon_{1}}})
    \end{split}
\end{equation}
where $^{j}\epsilon_{1}$, $^{j}\epsilon_{2}$, $^{j}a_{1}$, $^{j}a_{2}$ and $^{j}a_{3}$ are the parameters of $SQ_j$.
\vspace{-1mm}

According to Eq. \ref{eqn: HOCBF}, $h_{co}$ becomes positive when the proxy of $SQ_i$ is located outside the $SQ_j$; namely, two SQs do not collide. Thus, $h_{co}$ can be a CBF for the collision avoidance.
\vspace{-1mm}

\subsection{Outer-Loop Control Law}
\vspace{-1mm}

Given the target pose of the AM from Eq. \ref{target pose calculation}, the desired pose and twist, $[\boldsymbol{q_d};\boldsymbol{\theta_d}]$ and $[\boldsymbol{\dot{q}_d};\boldsymbol{\dot{\theta}_d}]$, need to be generated based on the two safety conditions. From Eq. \ref{eqn: lower and upper limits of propeller thrusts} (thrust limits), we can find $\boldsymbol{A_{\underline{T}}}, \boldsymbol{A_{\bar{T}}} \in {\mathbb{R}}^{6\times6}$ and $\boldsymbol{b_{\underline{T}}}, \boldsymbol{b_{\bar{T}}} \in {\mathbb{R}}^{6}$ s.t.
\begin{equation} \label{eqn: final inequality for thrust limits}
    \begin{split}
        \boldsymbol{A_{\underline{T}}}\boldsymbol{\dot{q}_d} \leq \boldsymbol{b_{\underline{T}}}, \ \boldsymbol{A_{\bar{T}}}\boldsymbol{\dot{q}_d} \leq \boldsymbol{b_{\bar{T}}}, 
    \end{split}
\end{equation}
Meanwhile, since $h_{co}$ is the function of $(\boldsymbol{q}, \boldsymbol{\theta})$, we need to differentiate it twice to attain the thrust inputs. Hence, Eq. \ref{eqn: HOCBF} becomes a high-order CBF (\cite{xiao2021high}), and the collision avoidance condition is formulated as follows:
 \begin{equation} \label{eqn: HOCBF condition}
     \ddot{h}_{co} + 2\alpha_{co}\dot{h}_{co} + \alpha^2_{co}h_{co} \geq 0
 \end{equation}
 where $\alpha_{co} > 0$ is a user-defined parameter. With the assumption that $\boldsymbol{\gamma_{3D,i}}$ and $\boldsymbol{\gamma_{3D,j}}$ slowly change with time, the time-derivatives of $h_{co}$ depend on those of $\boldsymbol{p_{i}}$ and $\boldsymbol{R_{i}}$. Therefore, $\ddot{h}_{co}$ and $\dot{h}_{co}$ can be expressed as follows:
\begin{equation*}
    \begin{split}
        \dot{h}_{co} =& \tfrac{\partial h_{co}}{\partial \boldsymbol{\Delta X}}\boldsymbol{\dot{(\Delta X)}} \\
        \ddot{h}_{co} =& \tfrac{\partial h_{co}}{\partial \boldsymbol{\Delta X}}\boldsymbol{\ddot{(\Delta X)}} + \boldsymbol{\dot{(\Delta X)}^{\top}} \tfrac{\partial^2 h_{co}}{\partial (\boldsymbol{\Delta X})^2} \boldsymbol{\dot{(\Delta X)}}.
    \end{split}
\end{equation*}
where we can find matrices $\boldsymbol{A_{\Delta X}} \in {\mathbb{R}}^{3\times9}$ and $\boldsymbol{b_{\Delta X}} \in {\mathbb{R}}^{3}$ such that $\boldsymbol{\ddot{(\Delta X)}} = \boldsymbol{A_{\Delta X}}[\boldsymbol{\ddot{q}};\boldsymbol{\ddot{\theta}}] + \boldsymbol{b_{\Delta X}}$ from Eq. \ref{eqn: Delta X}. Also, by substituting Eq. \ref{eqn: DOB-based controller} for Eq. \ref{eqn: rearranged dynamic model of the aerial manipulator}, $\boldsymbol{\ddot{q}}$ is rearranged as follows:
\begin{equation}
    \begin{split}
        \boldsymbol{\ddot{q}} = \boldsymbol{K_d}\boldsymbol{\dot{q}_d} - \boldsymbol{K_d}\boldsymbol{\dot{q}} + \boldsymbol{K_p}\boldsymbol{e} - \boldsymbol{\hat{M}^{-1}}(\boldsymbol{\phi})(\boldsymbol{\hat{d}} - \boldsymbol{d}).
    \end{split}
\end{equation}
Assume that $\boldsymbol{\ddot{\theta}} \approx \boldsymbol{\ddot{\theta}_d}$, then Eq. \ref{eqn: HOCBF condition} is rearranged as follows:
\begin{equation}
    \boldsymbol{A_{co}}[\boldsymbol{\dot{q}_d};\boldsymbol{\ddot{\theta}_d}] + d_{co} \leq b_{co}
\end{equation}
where $\boldsymbol{A_{co}}$ $\triangleq$ $-\tfrac{\partial h_{co}}{\partial \boldsymbol{\Delta X}}\boldsymbol{A_{\Delta X}}$ $\begin{bmatrix}
            \boldsymbol{K_d} & \boldsymbol{0_{6 \times 3}} \\
            \boldsymbol{0_{3 \times 6}} & \boldsymbol{I_3}
        \end{bmatrix}$, $d_{co}$ $\triangleq$ $\tfrac{\partial h_{co}}{\partial \boldsymbol{\Delta X}}\boldsymbol{A_{\Delta X}}$ $\begin{bmatrix}
            \boldsymbol{\hat{M}^{-1}}(\boldsymbol{\phi})(\boldsymbol{\hat{d}} - \boldsymbol{d}) \\
            \boldsymbol{0_{3 \times 1}}
        \end{bmatrix}$ and $b_{co}$ $\triangleq$ $\tfrac{\partial h_{co}}{\partial \boldsymbol{\Delta X}}$ $\boldsymbol{A_{\Delta X}}\begin{bmatrix}
            -\boldsymbol{K_d}\boldsymbol{\dot{q}} + \boldsymbol{K_p}\boldsymbol{e} \\
            \boldsymbol{0_{3 \times 1}} 
        \end{bmatrix}$ $+ \tfrac{\partial h_{co}}{\partial \boldsymbol{\Delta X}}\boldsymbol{b_{\Delta X}}$ $+ \boldsymbol{\dot{(\Delta X)}^{\top}} \tfrac{\partial^2 h_{co}}{\partial (\boldsymbol{\Delta X})^2} \boldsymbol{\dot{(\Delta X)}}$ $+ 2\alpha_{co}\tfrac{\partial h_{co}}{\partial \boldsymbol{\Delta X}}\boldsymbol{\dot{(\Delta X)}} + \alpha^2_{co}h_{co}$. 
As $d_{co}$ contains the disturbance estimation error, with $|d_{co}| \leq \sigma_{co}$, a more conservative collision avoidance constraint is expressed as:
\begin{equation} \label{eqn: final inequality for collision avoidance}
    \boldsymbol{A_{co}}[\boldsymbol{\dot{q}_d};\boldsymbol{\ddot{\theta}_d}] + \sigma_{co} \leq b_{co}.
\end{equation}
Finally, a quadratic programming (QP)-based outer-loop controller is constructed as follows:
\begin{equation} \label{eqn: outer-loop controller}
    \begin{split}
        \underset{[\boldsymbol{\dot{q}_d};\boldsymbol{\ddot{\theta}_d}]}{\textrm{min}} & \ \|\boldsymbol{\dot{q}_d} - \boldsymbol{\dot{q}_{d,ref}}\|^2_{\boldsymbol{Q_{\dot{q}}}} + \|\boldsymbol{\ddot{\theta}_d} - \boldsymbol{\ddot{\theta}_{d,ref}}\|^2_{\boldsymbol{Q_{\ddot{\theta}}}} \\
        \textrm{s.t. } & \textrm{Eq. \ref{eqn: final inequality for thrust limits} and Eq. \ref{eqn: final inequality for collision avoidance}}
    \end{split}
\end{equation}
where $\boldsymbol{\dot{q}_{d,ref}} \triangleq \boldsymbol{\Gamma_{q}}(\boldsymbol{q_t} - \boldsymbol{q_d})$ and $\boldsymbol{\ddot{\theta}_{d,ref}} \triangleq -2\boldsymbol{\Gamma_{\theta}}\boldsymbol{\dot{\theta}_d} + \boldsymbol{\Gamma^2_{\theta}}(\boldsymbol{\theta_t} - \boldsymbol{\theta_d})$ with weight matrices $\boldsymbol{Q_{\dot{q}}}, \boldsymbol{Q_{\ddot{\theta}}} \in \mathbb{R}^{3\times3}_{>0}$, and target-following coefficients $\boldsymbol{\Gamma_{q}}, \boldsymbol{\Gamma_{\theta}} \in \mathbb{R}^{3\times3}_{>0}$.
\vspace{-1mm}

\section{Simulation Results}
\vspace{-1mm}

We evaluate the proposed planner in two representative cluttered scenarios and compare it with baseline planners.
\vspace{-1mm}

\subsection{Results from whole-body motion planner}
\vspace{-1mm}

\begin{figure}[h!]  
\centering
	\includegraphics[width=0.875\columnwidth]{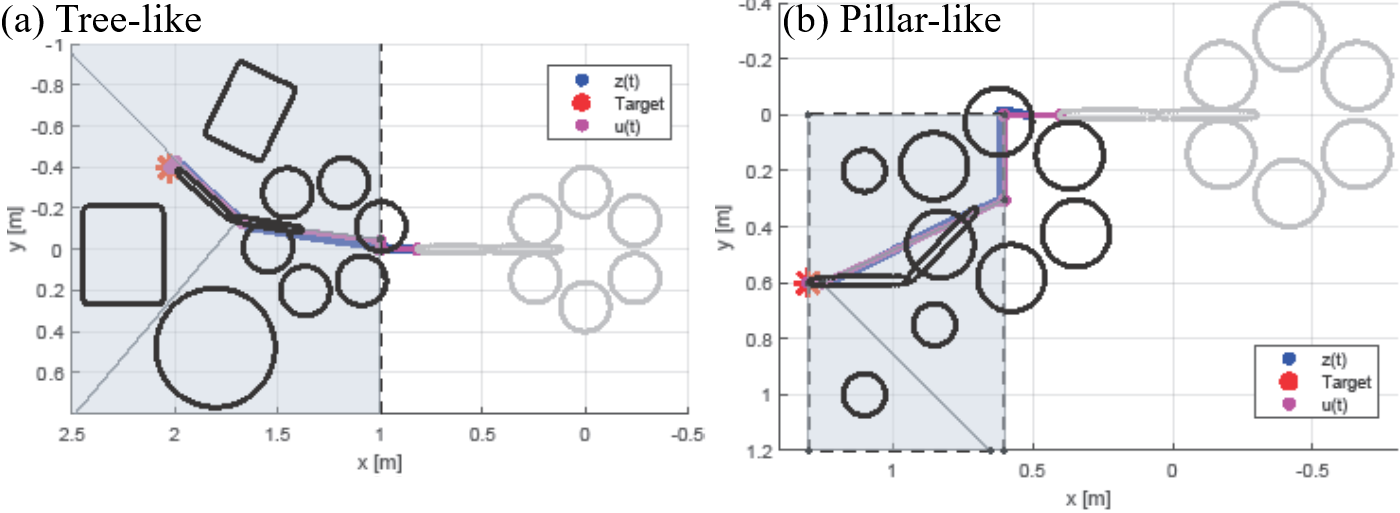}
  \vspace{-0.47cm}
\caption{Whole-body planning results in tree- and pillar-like environments. Blue polygons denote Voronoi regions, grey lines denote Voronoi boundaries, red the target, magenta $\V u(s)$, and blue the planned trajectory $\V z(s)$.}
\label{FIG:Voro}
\end{figure}

Fig. \ref{FIG:Voro} shows the planning results in tree-like and pillar-like environments. In both cases, the target is reachable only by exploiting the thin arm linkages, highlighting the necessity of whole-body planning.
\vspace{-1mm}


\vspace{-1mm}

\subsection{Benchmark}
\vspace{-1mm}

We compare the proposed planner with RRT* \cite{karaman2011sampling}, MPPI \cite{williams2017information}, and an ellipse-based ablation that replaces SQ obstacles with ellipses. RRT* is sampled in $\V z_{sys}$ with geometric collision checking, while MPPI uses $N=50$ rollouts and penalizes target distance and trajectory smoothness. The ablation uses the same planner settings as ours except for the obstacle representation.
\vspace{-1mm}

\begin{figure}[h!]
	\centering
	\includegraphics[width=0.58\columnwidth]{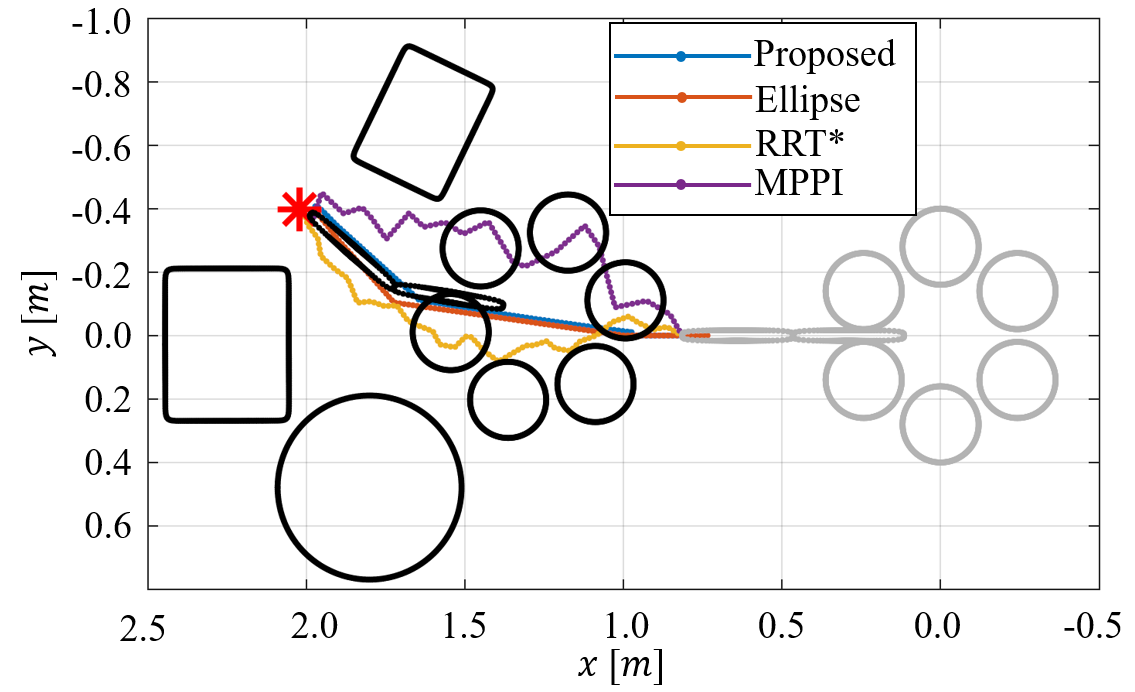}
    \vspace{-0.38cm}
	\caption{Comparison of end-effector trajectories by different planners: (a) the proposed SQ-based planner, (b) ablation study (ellipses) (c) RRT*, (d) MPPI.}
	\label{FIG:benchmark}
\end{figure}
\vspace{-1mm}

\begin{table}[h!]
\centering
\caption{Benchmark with related planners}
\vspace{-0.35cm}
\begin{tabular}{|c || c | c| c| c| c|} 
\hline
Method & Time & Min-distance & Arc-length & Jerkiness \\
\hline
Proposed & 0.128 & 0.0644 & 1.375 & 0.0016  \\
\hline
Ellipse & 0.113 & -0.0049 & 1.383 & 0.0001  \\
\hline
RRT* & 105.8 & 0.0314 & 1.565 & 0.0032 \\
\hline
MPPI & 0.552 & 0.0039 & 1.721 & 0.0057  \\
\hline
\end{tabular}
\label{table:benchmark}
\end{table}
\vspace{-1mm}

Fig. \ref{FIG:benchmark} and Table \ref{table:benchmark} report four metrics: online planning time, minimum robot-obstacle distance, trajectory arc-length, and jerkiness. The proposed method achieves the largest minimum distance, indicating better clearance than the baselines. Compared with the ellipse-based ablation, the SQ representation avoids excessive conservativeness and more accurately captures obstacle geometry. RRT* generates collision-free paths but requires significantly longer computation time, while MPPI has higher online cost and produces longer, less smooth trajectories. Overall, the proposed planner provides a favorable balance between safety, efficiency, and smoothness.
\vspace{-1mm}

\section{Experimental Results}
\vspace{-1mm}

\begin{figure}
    \centering
    \includegraphics[width=0.40\textwidth]{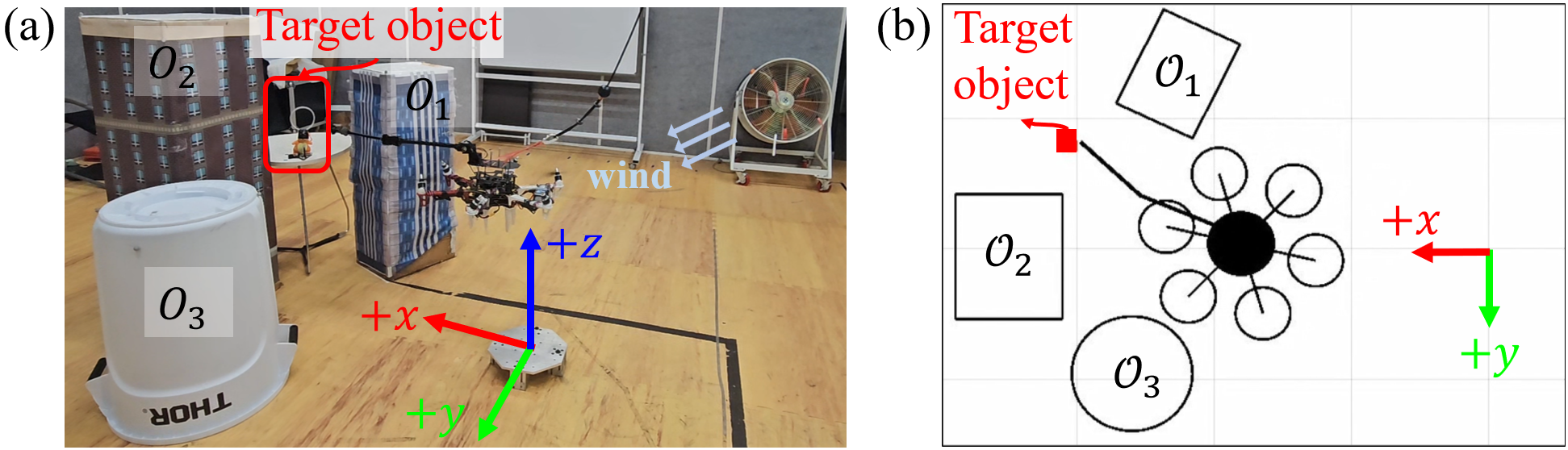}
    \vspace{-0.35cm}
    \caption{Experiment setup: (a) 3D side view in the laboratory environment. (b) 2D top view illustration. \textbf{}}
    \label{fig: experimental setting}
\end{figure}

We also validate our planner and controller with an actual object-picking experiment as in Fig. \ref{fig: experimental setting}.
\vspace{-1mm}

\vspace{-1mm}

\subsection{Experimental Setups and Scenario}
\vspace{-1mm}

Our AM combines a fully actuated hexacopter and a 3-DOF robot arm. The 3.5 kg hexacopter, based on a DJI F550 frame, uses six 9-inch APC LPB09045 propellers, KDE2314XF-965 motors with KDEXF-UAS35 ESCs, and custom 3D-printed thrust-tilting mounts. Power is supplied by separate 4S (Intel NUC) and 6S (ESCs) LiPo batteries. The arm uses three ROBOTIS XM430 actuators via a U2D2 interface. An Intel NUC running Ubuntu 20.04 and ROS Noetic handles control and OptiTrack-based navigation, while ESCs are governed by a Pixhawk 4. An industrial fan is utilized to simulate wind gusts, and parameter values are listed in Table \ref{table: planner parameters}.
\vspace{-1mm}


\begin{table}[h!]
\centering
\caption{Parameters for planner and controller}
\vspace{-0.35cm}
\begin{tabular}{|c || c |} 
\hline
$\alpha$, $T_d$, $d_0$ & 20 [-], $30$ s, $1.0$ mm \\
\hline
$(k_{\textrm{min}}, k_{\textrm{max}})$ & $(1.0\times10^{-7}, 1.0\times10^{3})$ N/m \\
\hline
$\boldsymbol{K_{tgt}}$ & 800diag$\{2.0, 2.0, 1.0\}$ [-]  \\ 
\hline
$\boldsymbol{a_0}$, $\boldsymbol{a_1}$, $\boldsymbol{\varepsilon_{dob}}$ & $\boldsymbol{I_6}$, $2\boldsymbol{I_6}$, $0.95\boldsymbol{I_6}$ [-] \\
\hline
$\alpha_{co}$, $\sigma_{co}$, $\underline{T}$, $\bar{T}$ & 5.0 [-], 1.0 [-], 1.0 N, 15 N \\ 
\hline
$\boldsymbol{Q_q}$, $\boldsymbol{Q_{\ddot{\theta}}}$, $\boldsymbol{{\Gamma}_q}$, $\boldsymbol{\Gamma_{\theta}}$ & $\textrm{blkdiag}\{\boldsymbol{I_3}, 3\boldsymbol{I_3}\}$, $4\boldsymbol{I_3}$, $4\boldsymbol{I_6}$, $5\boldsymbol{I_3}$ [-] \\
\hline
$\boldsymbol{K_p}$ & {blkdiag}\{6$\boldsymbol{I_2}$,8.0,80$\boldsymbol{I_2}$,35\} [-] \\
\hline
$\boldsymbol{K_d}$ & {blkdiag}\{5$\boldsymbol{I_2}$,6.0,35$\boldsymbol{I_2}$,20\} [-] \\
\hline
\end{tabular}
\label{table: planner parameters}
\end{table}
\vspace{-1mm}


As illustrated in Fig. \ref{fig: experimental setting}(a), obstacle 1 is a blue cuboid (0.4 $\times$ 0.31 $\times$ 1.03 $m^3$), obstacle 2 is a brown cuboid (0.48 $\times$ 0.39 $\times$ 1.27 $m^3$), and obstacle 3 is a white cylinder (radius 0.29 m, height 0.72 m). The poses of all obstacles and the target object are measured using the OptiTrack.
\vspace{-1mm}

The vehicle takes off and deploys the robot arm to its fully extended configuration ($\boldsymbol{\theta} = \boldsymbol{0_{3\times1}}$). Subsequently, the path planning algorithm is activated, after which the vehicle follows the generated trajectory, guides the end-effector through the ring affixed to the target, ascends to grasp it, and returns to the initial position.
\vspace{-1mm}


\subsection{Result and Discussion}
\vspace{-1mm}
\begin{figure}[!t]
    \centering
    \includegraphics[width=0.36\textwidth]{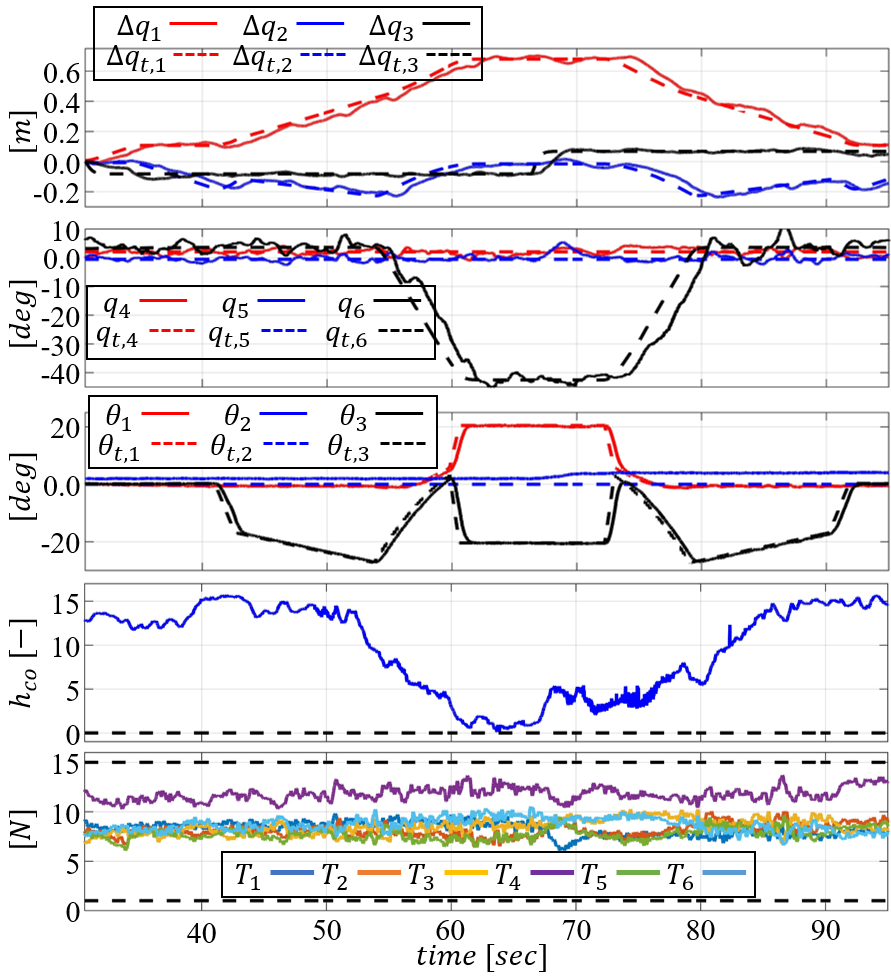}
    \vspace{-0.37cm}
    \caption{Histories of the multirotor's measured and target displacements from initial position ($\Delta q_i \triangleq [q_i - q_i(t_0)$ and $\Delta q_{t,i} \triangleq q_{t,i} - q_i(t_0)$, $(i = 1,2,3)$ and its measured and target Euler angles ($[q_4;q_5;q_6]$ and $[q_{t,4};q_{t,5};q_{t,6}]$), robot arm's measured and target joint angles ($\boldsymbol{\theta}$ and $\boldsymbol{\theta_t}$), collision avoidance CBF, $h_{co}$, and thrust values during approaching, picking, and returning.}
    \label{fig: whole_history}
\end{figure}
\vspace{-1mm}

\begin{figure}[!t]
    \centering
    \includegraphics[width=0.31\textwidth]{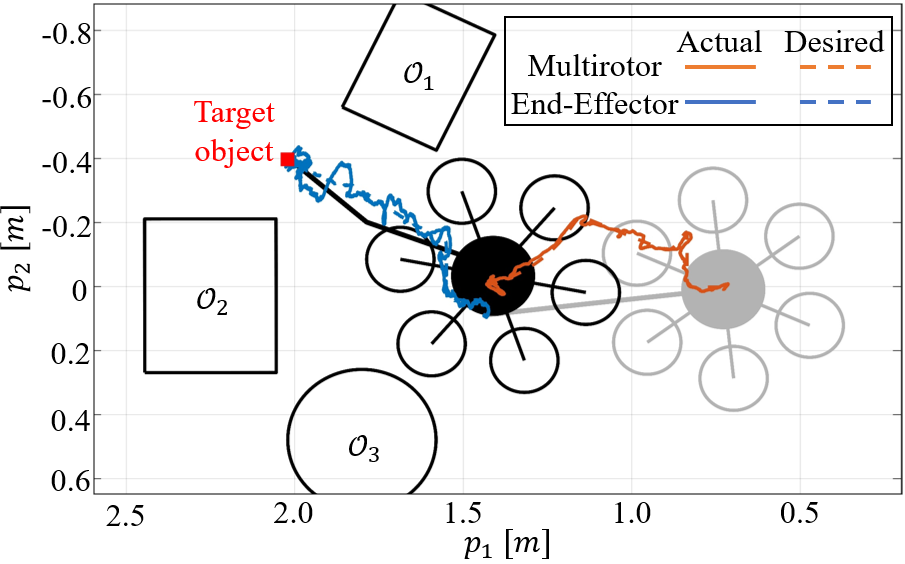}
    \vspace{-0.4cm}
    \caption{2D top view trajectories of the multirotor and the end-effector from the initial to picking poses.}
    \label{fig: whole_trajectory}
\end{figure}

Figs. \ref{fig: whole_history} and \ref{fig: whole_trajectory} indicate that the task is completed without collisions. In Fig. \ref{fig: whole_history}, $h_{co}$ remains positive throughout the operation, confirming collision-free performance despite tracking errors in the multirotor’s position and attitude, and all motor thrusts adhere to their limits. These results verify the effectiveness of our controller for instantaneous collision avoidance. Fig. \ref{fig: whole_trajectory} displays the multirotor navigating and rotating within a narrow path, while the robot arm simultaneously bends to reach the target object. The attached video (\url{https://youtu.be/hQYKwrWf1Ak}) presents a more detailed visualization of the experimental process.
\vspace{-1mm}

\vspace{-1mm}

\section{Conclusion}
\vspace{-1mm}

This paper proposes whole-body motion planning and safety-critical control for an AM using superquadrics and proxies, where both the multirotor platform and the robot arm linkages independently avoid collisions with surrounding obstacles. First, we derive the dynamic model of the AM by attaching a multi-DOF robot arm to a fully actuated hexarotor. Second, we design a maximum-clearance whole-body motion planner by integrating the SQ–proxy representation with Voronoi diagrams on an equilibrium manifold. Third, a controller is developed to ensure thrust limits and collision avoidance. Through comparative simulations, we show that our framework outperforms sampling-based planners in cluttered environments, achieving faster, safer, and smoother trajectories, and surpasses ellipsoid-based ablation methods in accurately representing geometry. Additionally, we validate the dynamic feasibility and robustness of our approach on a physical AM. For future work, both planning and control methods will be extended to 3D environments and experimentally validated in more cluttered scenarios that include obstacles only allowing narrow gaps near the target location.
\vspace{-1mm}

\bibliography{ifacconf}             
\end{document}